\documentclass[letterpaper, 10 pt, conference]{ieeeconf}
\usepackage{cite} 
\usepackage{multirow}
\usepackage[utf8]{inputenc}
\usepackage[misc]{ifsym}
\usepackage{graphicx}
\usepackage{tikz}
\usepackage{algorithm}
\usepackage{algorithmic}
\usepackage{pifont}
\newcommand{\cmark}{\ding{51}}%
\newcommand{\xmark}{\ding{53}}%
\usepackage[misc]{ifsym}
\usepackage{tikz}
\usepackage{pdfpages}
\usepackage{graphicx}
\usepackage{amsmath}
\usepackage[utf8]{inputenc}
\usetikzlibrary{arrows,shapes,positioning,shadows,trees,calc,decorations.markings}

\usepackage{amssymb}
\usepackage{hyperref}

\makeatletter
\newcommand{\printfnsymbol}[1]{%
  \textsuperscript{\@fnsymbol{#1}}%
}
\makeatother

\IEEEoverridecommandlockouts                

\title{\LARGE \bf
Learning Domain Adaptation with Model Calibration for Surgical Report Generation in Robotic Surgery
}

\author{Mengya Xu\printfnsymbol{1}\thanks{\printfnsymbol{1} equal contribution}, Mobarakol Islam\printfnsymbol{1}, Chwee Ming Lim  and Hongliang Ren

\thanks{M. Xu and H. Ren are with Dept. of Biomedical Engineering, National University of Singapore, Singapore and NUSRI Suzhou China; Corresponding author: Hongliang Ren, hlren@ieee.org http://labren.org}
\thanks{H. Ren is with Dept. of Electronic Engineering, The Chinese University of Hong Kong} 
\thanks{M. Islam is with Dept. of Computing, Imperial College London, UK and was with Dept. of Biomedical Engineering, National University of Singapore, Singapore}
\thanks{C. M. Lim is with Dept. of Otolaryngology-Head and Neck Surgery, Duke-NUS Medical School, Singapore}
}

\usepackage{amsmath}
\begin{document}

\maketitle
\thispagestyle{empty}
\pagestyle{empty}

\begin{abstract}

Generating a surgical report in robot-assisted surgery, in the form of natural language expression of surgical scene understanding, can play a significant role in document entry tasks, surgical training, and post-operative analysis. Despite the state-of-the-art accuracy of the deep learning algorithm, the deployment performance often drops when applied to the Target Domain (TD) data. For this purpose, we develop a multi-layer transformer-based model with the gradient reversal adversarial learning to generate a caption for the multi-domain surgical images that can describe the semantic relationship between instruments and surgical Region of Interest (ROI). In the gradient reversal adversarial learning scheme, the gradient multiplies with a negative constant and updates adversarially in backward propagation, discriminating between the source and target domains and emerging domain-invariant features. We also investigate model calibration with label smoothing technique and the effect of a well-calibrated model for the penultimate layer's feature representation and Domain Adaptation (DA). We annotate two robotic surgery datasets of MICCAI robotic scene segmentation and Transoral Robotic Surgery (TORS) with the captions of procedures and empirically show that our proposed method improves the performance in both source and target domain surgical reports generation in the manners of unsupervised, zero-shot, one-shot, and few-shot learning.

\end{abstract}

\section{Introduction} 
The popularity of Minimally Invasive Surgeries (MIS) in modern medical treatment has brought new opportunities and challenges for automated surgical scene understanding, which can be used to empower Computer Assisted Interventions (CAI) and is the foundation of intelligent systems, such as robotic vision, surgical training applications, and medical report generation. This kind of intelligent system can remind surgeons not to miss important steps during complex surgery. By integrating with natural language report generation, the intelligent system can also eliminate the frustrating task of document entry task for clinicians, allowing them to focus on patient-centric activities. It can free doctors and nurses from the low-value task of writing reports of surgical procedures that are more suitable for machines. Besides, such a natural language record of the surgical procedure has a detailed post-operative analysis of the surgical intervention.

Automatically generating the description for a given surgical procedure is a complicated problem since it requires an algorithm to complete several computer vision and Natural Language Processing (NLP) tasks, such as object recognition, relationships understanding between vision and text elements, and then generate a sequence of words. Existing studies for automatic report generation focus on medical images, such as radiology and pathology images. In this case, the report is a diagnosis report which will describe whether the body part examined in the imaging technique was normal, abnormal, or potentially abnormal~\cite{jing2017automatic}. In our work, the automatically generated report focuses more on understanding and describing the surgical scene. It describes which surgical instrument appears in the surgical scene and the instrument-tissue interaction.

Existing image captioning models cannot generalize well to the Target Domain (TD) images. Learning a predictor when there is a shift in data distribution between training and validation is considered as Domain Adaptation (DA). The cost of generating labels for data is high, which often becomes one of the biggest obstacles preventing the application of the machine learning approach. However, the model with the DA ability can work when the TD lacks labels.

\begin{figure*}[!hbpt]
\centering
\includegraphics[width=0.75\linewidth]{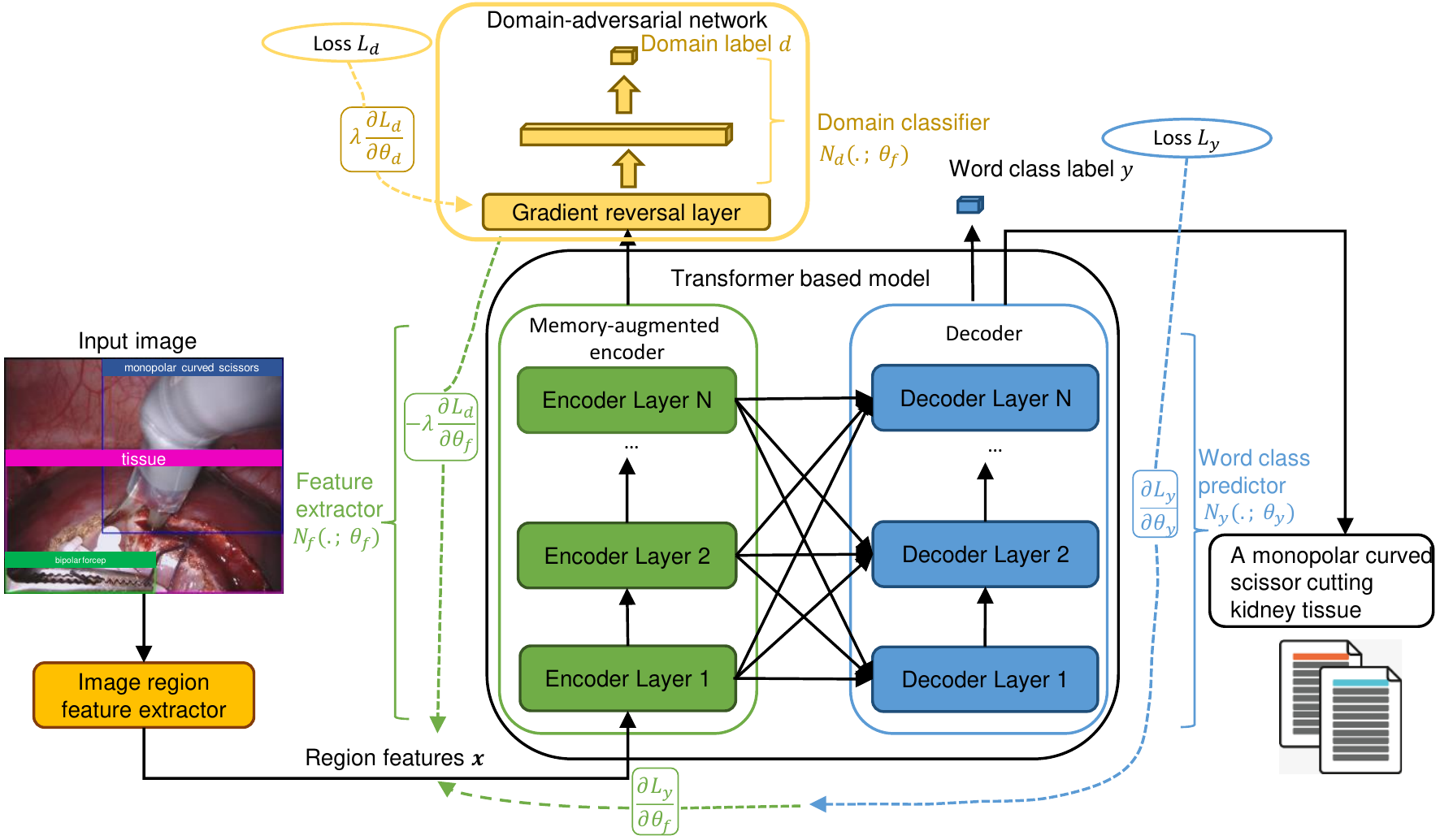}
\caption{The architecture of the proposed network. The network comprises a region feature extractor, multi-layer Transformer-like encoder-decoder, and a domain-adversarial network. ResNet18 is used to extract region features in input images. The domain-adversarial network, which consists of a GRL and a domain classifier, is connected with the last encoding layer to distinguish whether the sample is from the SD or the TD. The image encoder incorporated with domain-adversarial neural network takes in the region features and learns the domain-invariant features with an adversarial learning scheme and attempts to understand relationships between regions. The language decoder reads the output of each encoding layer to generate the output sentence word by word. The stacked sentence can form a medical report.}

\label{fig:Architecture of the network}
\end{figure*}
 
\subsection{Related Work}

\subsubsection{Surgical Scene Understanding}
Surgical scene understanding is significant for image-guided robot-assisted surgery. Incorporating scene understanding capability in robotic surgery provides possibilities for future semi-automated or fully automated operations. Our previous works include surgical instrument tracking with task-directed attention~\cite{islam2020ap}~\cite{islam2019learning}, and real-time instrument segmentation ~\cite{islam2019real}. After the surgical instrument recognition problem has been solved to a large extent, the next goal is to get out the instrument itself and focus on the relationships between the instruments and  the Region of Interest (ROI). Therefore, a graph-based network is introduced in~\cite{islam2020learning} to do deep reasoning for the surgical scene, learn to infer a graph structure of instruments-tissue interaction, and predict the relationship between instruments and tissue. The further intention is that we want to express this prediction of their relationship in natural language, which can have richer scene information and the ability to interact with people. 

\subsubsection{Transformer based model from language to vision}
The Transformer~\cite{vaswani2017attention} is proposed for language translation which achieves the state-of-the-art performance. In combining scene understanding and natural language, Convolutional Neural Network (CNN) is used to extract visual features, and Recurrent Neural Network (RNN) is widely adopted as the language models, which can be summarized as "CNN + RNN" framework~\cite{shi2020improving}. Although this framework is widely adopted, the RNN-based model is severely limited by its sequential nature and difficult to parallelize characteristics. Some recent researches have used the Transformer model as the language model for scene understanding in which the input sequence can be passed in parallel.~\cite{herdade2019image} still utilized the original architecture of the Transformer and explored the geometric relationships between detected objects. A memory-augmented recurrent Transformer~\cite{lei2020mart} is employed to do video captioning.

\subsubsection{Label Smoothing}
Label Smoothing (LS) is a regularization technique that flattens the true label with a uniform distribution during cross-entropy loss calculation in network training \cite{szegedy2016rethinking}. Recently, there is evidence that LS can prevent over-confidence in prediction and improve the model calibration~\cite{muller2019does}. Moreover, LS is also used to enhance the prediction uncertainty \cite{zhang2020self} and the teacher-free knowledge distillation \cite{yuan2019revisit}. In our previous work, we present that LS can improve the feature representation in the penultimate layer and effective as an object feature extraction model \cite{islam2020learning}.

\subsubsection{Domain Adaptation}
 The model's performance tends to drop when evaluated on a TD dataset, which is different from the dataset used for training. This limitation motivates the research on DA~\cite{valverde2019one} in multiple sclerosis lesion segmentation in one shot way. The method for DA in~\cite{orbes2019multi} consists of domain adversarial learning and consistency training. Many methods attempt to match the feature space distribution in the Source Domain (SD) and the TD for unsupervised DA (UDA). To this end,~\cite{borgwardt2006integrating} reweigh or select examples of the SD.~\cite{ganin2016domain} achieves this matching by modifying and changing the feature representation itself. For supervised DA, the approaches utilize the labeled data from the TD to "fine-tune" the network trained on the SD. In our work, the unsupervised and semi-supervised DA are explored.

\subsection{Contributions}
 Our contributions are summarized as follows:
\begin{itemize}
    \item[--] Propose transformer-based multi-layer encoder-decoder architecture with the gradient reversal adversarial learning scheme to generate the surgical report in robotic surgery. 
    \item[--] Investigate model calibration with label smoothing cross-entropy loss for feature extraction from penultimate layer and UDA.
    \item[--] Design domain-adaptation in various unsupervised and semi-supervised manners such as zero-shot, one-shot, and few-shot training.
    \item[--] Annotate robot-assisted surgical datasets with proper captions to generate the surgical report in two different surgical domain datasets of MICCAI robotic scene segmentation challenge and Transoral Robotic Surgery (TORS).

\end{itemize}

\section{Methods}

\subsection{Background}

Transformer-based multi-layer encoder-decoder in~\cite{cornia2020meshed} is a fully-attentive model and shows excellent performance for image captioning tasks. It is a variant of the original Transformer~\cite{vaswani2017attention} which achieves the state-of-the-art results in machine translation and language understanding. In~\cite{cornia2020meshed}, the encoder module takes regions from images as input and understands relationships between regions. The decoder reads each encoding layer's output to model a probability over words in the vocabulary and generates the output sentence word by word by feeding the predicted word back as input at the next time step. Compared with~\cite{vaswani2017attention},~\cite{cornia2020meshed} has made two changes: 1) multi-head attention in image encoder is augmented with memory to understand the semantic relationships between detected input objects; 2) the cross-attention in language decoder is devised to utilize all encoding layers, rather than attending only the last encoding layer. 

Despite the excellent accuracy of the deep learning frameworks in vision tasks, it performs poorly in the TD dataset. To solve it,~\cite{ganin2016domain} proposes a novel representation learning method called domain adversarial learning for DA. The goal is to learn features that have 1) distinction for the major task during learning on the SD and 2) no distinction regarding the change of domains. H-divergence measures the distance between the source and the target distributions used by~\cite{ben2007analysis}. A Gradient Reversal Layer (GRL) with an adversarial learning scheme is used to domain adaption tasks with classification problem \cite{ganin2016domain}.

Most recently, it is found that the current deep learning models are poorly calibrated, which drops the generalization capability of the model \cite{muller2019does, guo2017calibration, lakshminarayanan2017simple}. LS is showing evidence to improve model calibration and prevent the over-confidence of model learning.~\cite{muller2019does} presents that LS calibrates the representations learned by the penultimate layer and makes the features of the same class form a tight cluster.
 
Inspiring by the above works, we design a transformer-based multi-layer encoder-decoder architecture with a GRL domain classifier and label smoothing cross-entropy loss to generate the surgical reports in robotic surgery.

\subsection{GRL Domain Classifier}
Domain classifier forms of GRL with a discriminator and trains in an adversarial manner to learn domain invariant features \cite{ganin2016domain}. It trains with adversarial labels that 0 label for the SD and 1 label for the TD. GRL just acts as identity transformation layer, $G(x) = x$ in forwarding propagation and a negative constant multiplies the gradient in backward propagation ($\frac{dG}{dx} = -I$, where I is an identity matrix) and then passes it to the preceding layer. By this way, GRL subtracts the gradients of the main network (our caption generation model) and domain classifier instead of being summed during Stochastic Gradient Descent (SGD) training. On the other hand, this prevents SGD from making features dissimilar across the domains. In this case, the model attempts to learn domain invariant features, making the domain classifier indistinguishable.

\begin{figure}
\centering
\includegraphics[width=.7\linewidth]{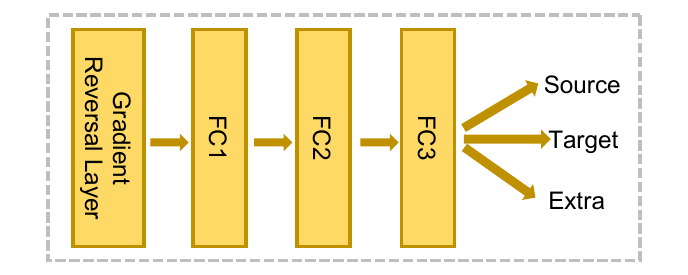}
\caption{Proposed GRL domain classifier. It consists of a GRL layer and three fully-connected layers. The output has 3 classes: SD, TD, and the extra one.}
\label{fig: novel_GRL_domain_classifier}
\end{figure}

We propose two significant improvements of the GRL domain classifier module: (1) we assign the discriminator with 3 classes (0 for source, 1 for target, and 2 for extra class) to enhance learning capacity, which will make the discriminator harder to fool during training; (2) integrate an additional fully-connected layer to boost feature learning, as shown in Fig. \ref{fig: novel_GRL_domain_classifier}

\subsection{LS for model calibration and DA}
A well-calibrated model estimates confidence probability, which represents true correctness of likelihood and leads to better generalization. LS, where a model is trained on a smoothened version of the true label with Cross-Entropy (CE) loss, shows great effectiveness in improving model calibration \cite{muller2019does}. If smoothened label, $T_{LS} = T(1-\epsilon) + \epsilon/K$ then CE loss with LS can be formulated as $CE_{LS} = - \sum_{k=1}^{K} T_{LS} log (P)$ where true label $T$, smoothing factor $\epsilon$, total number of classes $K$  and predicted probability $P$. A recent study \cite{islam2020learning} investigates that LS learns better feature representation in the penultimate layer, which is effective in object feature extraction. In this work, we observe the behavior of LS trained model in the DA task.

\subsection{Network Architecture}

The whole network consists of an image region feature extractor, Transformer-based multi-layer encoder-decoder~\cite{cornia2020meshed}, and a GRL domain classifier network. We choose a light-weight feature extraction model of ResNet18 \cite{he2016deep}, similar as \cite{islam2020learning}, to obtain feature vectors of surgical objects (details in section \ref{sec:feature_extraction}). The GRL domain classifier network consists of a discriminator followed by the GRL as in Fig. \ref{fig:Architecture of the network}. The encoder block forms of memory augmented self-attention layer and feed-forward layer and a decoder block consists of self-attention on words and cross attention over the outputs of all the encoder layers, similar to \cite{cornia2020meshed}. There are three encoder and decoder blocks stack to encode input features and predict the word class label. The output of the final encoder layer is fed to the proposed GRL domain classifier. Each training iteration calculates losses from both SD and TD, and then caption prediction loss is combined with domain classifier losses to update the model parameters in an adversarial manner. We utilize CE loss with LS as the caption prediction loss and vanilla CE as the domain loss function. Therefore, the model loss can be formulated as

\begin{equation}
\centering
\label{eqn:loss}
L = L_{y} + L_{d},
\end{equation}

where $L_{y} = CE_{LS}$ is the caption prediction loss which is CE loss with LS and $L_{d} = L_{S} + L_{T}$ is the domain loss which is the fused losses of the SD loss $L_{S}$ and the TD loss $L_{T}$. 

The GRL domain classifier leads to the appearance of features that have a distinction for the captioning task and domain-invariance at the same time. The model attempts to learn a representation allowing the decoder to predict the word class, also weakening the ability of the domain classifier.

\section{Experiments}

\subsection{Dataset} 
\subsubsection{Robotic scene segmentation challenge} 
The SD dataset is from the robotic instrument segmentation dataset of MICCAI endoscopic vision challenge 2018 \cite{allan20202018}. The training set includes 14 robotic nephrectomy operations obtained by the da Vinci X or Xi system. There are 149 frames in each video sequence, and the frame has a dimensionality of 1280*1024. A total of 9 objects appear in the dataset, including 8 Instruments: bipolar forceps, prograsp forceps, monopolar curved scissors, clip applier, suction, ultrasound probe, stapler, and large needle driver respectively and kidney tissue. These surgical instruments have a variety of different semantic relationships and interactions with the tissue. A total of 11 kinds of semantic relationships are identified to generate the natural language description for images. The identified semantic relationships include manipulating, grasping, retracting, cutting, cauterizing, looping, suctioning, clipping, ultrasound sensing, stapling, and suturing. The Sequences 1st, 5th, 16th are chosen for the validation, and the remaining 11 sequences are selected for the training following the previous work~\cite{islam2020learning}. The training and validation data sequences are carefully chosen to ensure that most interactions are presented in both sets. 

\subsubsection{TORS dataset}
We have collected the TD dataset of TORS surgery using the Da Vinci Robot on human patients. All 9 patients have oropharynx cancer. Among these 9 patients, there are 8 males and 1 female with ages ranging from 23 to 72 and multiple cancer sites like the tonsil, the base of the tongue, and the posterior pharyngeal wall. We filter out some poor-quality videos that remain still and are not undergoing surgery. A total of 181 frames were selected based on the requirement to have at least one same surgical instrument as the robotic scene segmentation challenge dataset. These 181 frames are cropped and then resized to the same shape with MICCAI Robotic scene segmentation challenge dataset. A total of 5 objects are in the dataset, including tissue, clip applier, suction, spatulated monopolar cautery, maryland dissector. The 5 kinds of relationships comprise manipulating, grasping, cauterizing, suctioning, clipping. We split the dataset based on different patients. The training dataset has 48 frames collected from Patient [1, 2, 3, 4, 5, 6, 7] and the validation dataset has 133 frames collected from Patient [8, 9].

\begin{figure}[!h]
\centering
\includegraphics[width=1\linewidth]{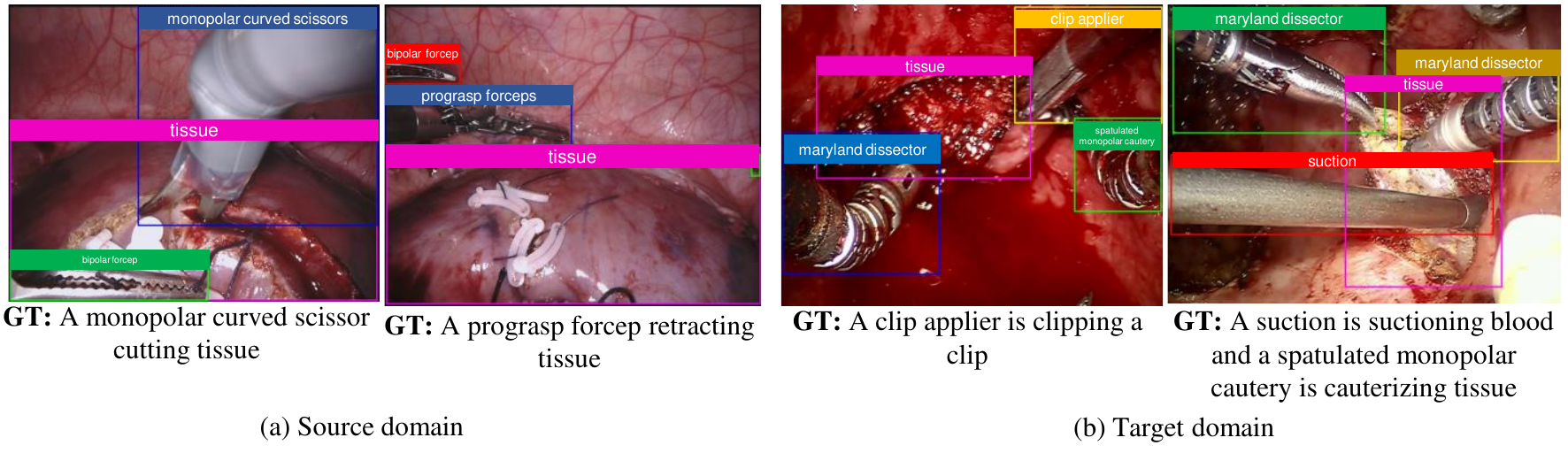}
\caption{Visualization of image-caption pairs from the SD and the TD. There are many unique ground truth captions}
\label{fig:dataset_visualization}
\end{figure}

\subsubsection{Domain Shift}

\begin{figure}[!hbpt]
\centering
\includegraphics[width=0.6\linewidth]{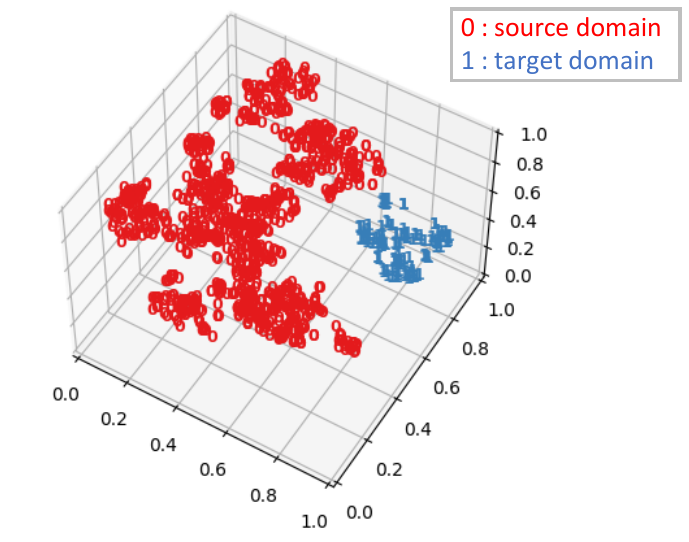}
\caption{t-SNE of samples before DA. The "0" indicated by red color are examples from the SD, and the "1" indicated by blue color are examples from the TD. The SD examples do not overlap with examples from the TD because of the distribution shift between these two domains.}
\label{fig:tSNE before domain adapatation}
\end{figure}

Fig. \ref{fig:dataset_visualization} shows example image-caption pairs from the SD and the TD. The SD is about robotic nephrectomy surgery, and the TD is about TORS. Compared with the SD, the TD has a different surgical background, and the surgery is performed on different tissue. Besides, the TD does not share all surgical instruments with the SD. Compared with the SD, the TD has two new instruments: spatulated monopolar cautery and maryland dissector. The distribution shift between the SD and TD can be distinguished from Fig. \ref{fig:tSNE before domain adapatation}, which shows the t-SNE of examples before DA from a three-dimensional (3D) perspective, which can prove the distribution shift between these two domains.

\begin{table*}
\centering
\caption{Performance of the proposed model on SD and TD dataset. Evaluation metrics such as BLEU-n, METEOR, ROUGR, and CIDEr are used to assess the similarity between the generated sentence and the ground-truth}
\begin{tabular}{c| c| c c c c c c c c c c}
\hline
\multicolumn{2}{c|}{}& & \textbf{BLEU-1}$\uparrow$ & \textbf{BLEU-2}$\uparrow$ & \textbf{BLEU-3}$\uparrow$ &\textbf{BLEU-4}$\uparrow$ & \textbf{METEOR}$\uparrow$ & \textbf{ROUGE}$\uparrow$ & \textbf{CIDEr}$\uparrow$ \\ \hline
\multicolumn{2}{c|}{\multirow{2}{*}{SD}} & $M^2$ Transformer \cite{cornia2020meshed} & 0.5054 & 0.4543 & 0.4055 & 0.3646 & 0.4441 & 0.6355 & 1.7878\\
\multicolumn{2}{c|}{} & Ours         & 0.5228 & 0.4730 & 0.4262 & 0.3861 & 0.4567 & 0.6495 & 2.2598\\
\hline
\multirow{8}{*}{TD}& \multirow{2}{*}{UDA} & $M^2$ Transformer \cite{cornia2020meshed} & 0.2302 & 0.1059 & 0.0469 & 0.0267 & 0.1286 & 0.2956 & 0.1305\\
& & Ours  & 0.2493 &  0.1150 & 0.0517 & 0.0289 & 0.1390 & 0.3129 & 0.1517\\
\cline{2-10}

& \multirow{2}{*}{Zero-shot} & $M^2$ Transformer \cite{cornia2020meshed} & 0.3204 & 0.2463 & 0.1923 & 0.1502 & 0.2371 & 0.4413 & 0.2874\\
& & Ours  & 0.3118 & 0.2406 & 0.185  & 0.1409 & 0.2401 & 0.4336 & 0.3395\\
\cline{2-10}

& \multirow{2}{*}{One-shot} & $M^2$ Transformer \cite{cornia2020meshed} & 0.3746 & 0.3285 & 0.2939 & 0.2646 & 0.3449 & 0.5101 & 0.6367\\
& & Ours  & 0.4042 & 0.372  & 0.3433 & 0.3161 & 0.4066 & 0.5385 & 0.8615\\
\cline{2-10}

& \multirow{2}{*}{Few-shot} & $M^2$ Transformer \cite{cornia2020meshed} & 0.4096 & 0.3803 & 0.3532 & 0.3265 & 0.4203 & 0.5489 & 0.9770\\
& & Ours & 0.4141 & 0.3888 & 0.3637 & 0.3375 & 0.4357 & 0.5538 & 0.9828\\

\hline

\end{tabular}
\label{table:full_performance}
\end{table*}

\subsubsection{Annotation}
The datasets are annotated with the appropriate caption by an experienced clinical expert in robotic surgery. The caption follows the description pattern of $\langle$ object1, predicate, object2 $\rangle$, which can concisely and accurately convey information of the surgical scene, for instance, "A monopolar curved scissor is cutting tissue."

\subsection{Implementation details}

\subsubsection{Image region feature extractor}
\label{sec:feature_extraction}
\begin{figure}[!hbpt]
\centering
\includegraphics[width=1\linewidth]{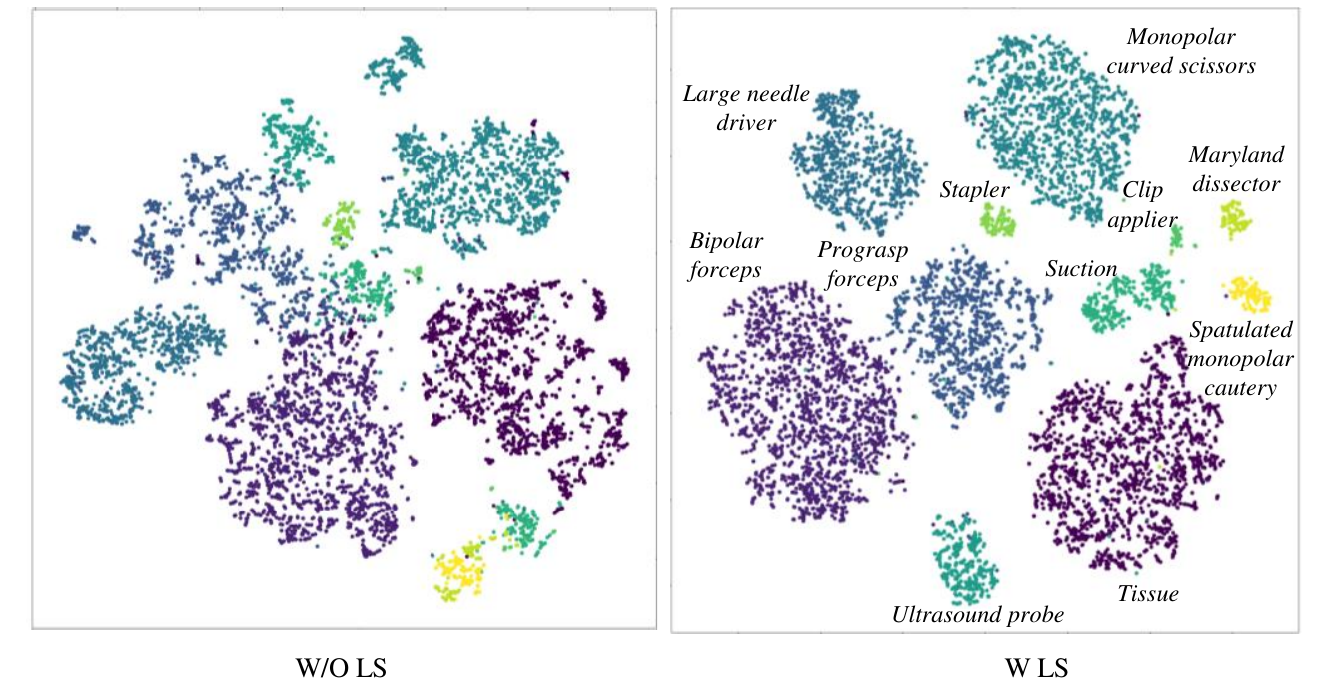}
\caption{The extracted features of the penultimate layer for ResNet18 w and w/o label smoothing (LS). The t-SNE plots show that the extracted features with LS are more distinguishable and tighter clusters for the same class.}
\label{fig:tsne_ls_features}
\end{figure}

We use the ResNet18 classification model to extract the feature vector for surgical objects. We crop the surgical instruments and ROI tissue and train the ResNet18 with CE loss combined with LS by following our previous work \cite{islam2020learning}. The penultimate layer's feature vector of size 512 is extracted for each surgical object.  Fig. \ref{fig:tsne_ls_features} shows the extracted features without (w/o) and with (w) LS. The features extracted by the LS model are more distinguishable among the classes.

\begin{figure*}[!hbpt]
\centering
\includegraphics[width=1\linewidth]{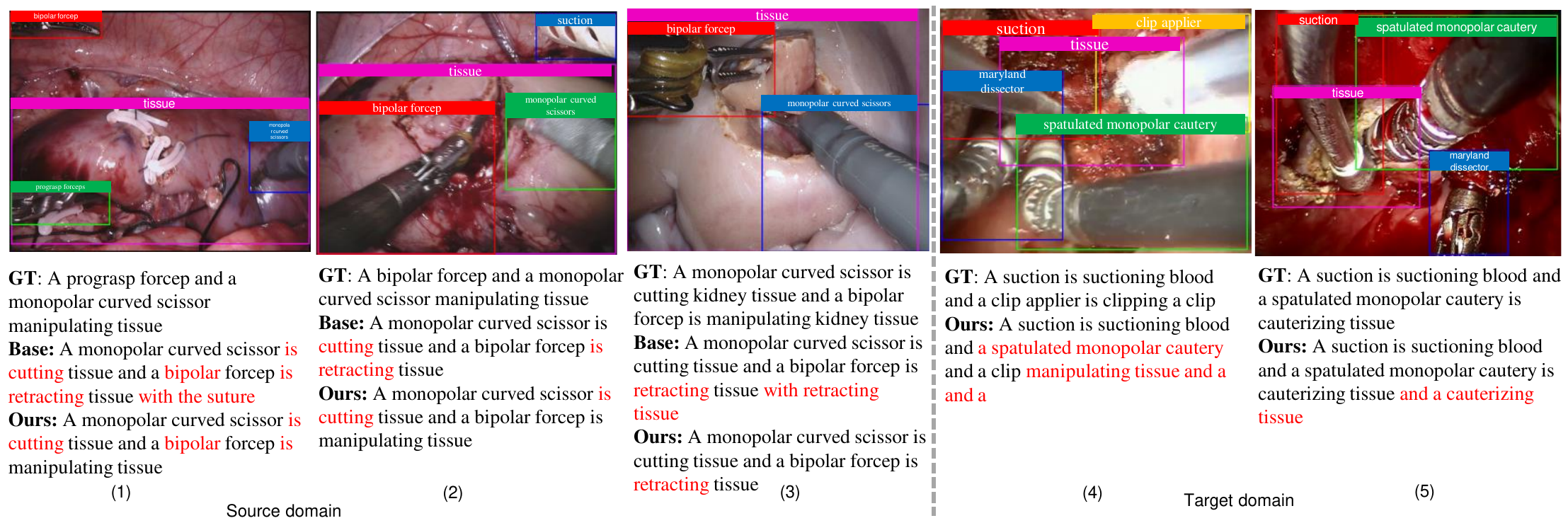}
\caption{Several image captioning examples generated by the base model and our model. Compared with the base model, the sentences predicted by our model are more reasonable and accurate.}
\label{fig:Results visualization}
\end{figure*}

\subsubsection{Vocabulary and tokenization}
All captions will go through a series of processing, including converting them to lower cases, removing punctuation characters, and tokenization using the spacy NLP toolkit. Vocabulary size of 49 is built based on both SD and TD, which include unique words in the captions and special tokens ($\langle$ unk $\rangle$, $\langle$ pad $\rangle$, $\langle$ bos $\rangle$, and $\langle$ eos $\rangle$). At test time, the model predicts the next word given the previously predicted words with beam search.

\subsubsection{Hyper-parameters}
Following the experimental setting in~\cite{cornia2020meshed}, when training with cross-entropy loss, the learning rate scheduling strategy of~\cite{vaswani2017attention} with a warmup equal to 10000 iterations is used. The number of epochs for training is set to 50. The seed is set to a fixed number to make the results reproducible. We train the model using Adam optimizer with a batch size of 50 and a beam size of 5. The network was implemented by PyTorch and trained in the NVIDIA RTX 2080 Ti GPU.

\section{Results and Evaluation}

\subsection{Evaluation Metrics}
We evaluate the approach using four commonly used metrics for image captioning, namely BLEU-n~\cite{papineni2002bleu}, ROUGE~\cite{lin2004rouge}, METEOR~\cite{banerjee2005meteor}, and CIDEr~\cite{vedantam2015cider}. BLEU-n measures the precision of n-grams between the ground-truth and the predicted sentences. Usually, n = 1, 2, 3, 4. 
To measure model calibration error, we use well-known metrics like Expected Calibration Error (ECE) \cite{guo2017calibration}, Static Calibration Error (SCE)\cite{nixon2019measuring}, Thresholded Adaptive Calibration Error (TACE), and Brier Score (BS) \cite{nixon2019measuring, ashukha2020pitfalls}.

\subsection{Caption generation with the SD dataset}
Table \ref{table:full_performance} represents the overall performance of the proposed approach with validation and domain adaption settings. The validation metrics are calculated with the SD dataset. Our method achieves better performance in almost all the metrics.

\subsection{Caption generation with the TD dataset}
Our model's domain adaption performance is evaluated with extensive unsupervised and semi-supervised settings such as zero-shot, one-shot, and few-shot training, as shown in Table \ref{table:full_performance}. All TD experiments share the same validation set and just use a different number of training images. In UDA, the model trained on SD is evaluated directly on TD. In zero-shot DA, the trained model is fine-tuned with the frames whose captions cover 85\% of the words in the vocabulary from the TD dataset. In one-shot DA, it is fine-tuned with images whose captions cover all the words with the smallest amount of images. Few images whose captions cover all the words are used to fine-tune the model. The proposed model has also outperformed the base model for these experiments.

\subsection{Model Calibration Vs. DA }

\begin{table}[!h]
\caption{Model miscalibration quantification for our model w and w/o LS. Evaluation metrics such as Expected Calibration Error (ECE), Static Calibration Error (SCE), Thresholded Adaptive Calibration Error (TACE), Brier Score (BS) are used to calculate the calibration error for each model.}
\label{tab:calibration}
\centering
\begin{tabular}{c|c|c|c|c|c}
\hline
\textbf{Ours} &\textbf{TD(UDA)} &\multicolumn{4}{c}{\textbf{Calibration Error}} \\ \hline
\textbf{Methods} & \textbf{BLEU-1}$\uparrow$ &\textbf{ECE}$\downarrow$ & \textbf{SCE}$\downarrow$ & \textbf{TACE}$\downarrow$ & \textbf{BS}$\downarrow$\\ \hline 
w LS &0.2493 &0.1768 &0.0493 &0.0489 &0.5063\\ \hline
w/o LS &0.2302 &0.2001 &0.0533 &0.0531 &0.9585\\ \hline

\hline
\end{tabular}
\label{table:calibration_vs_da}
\end{table}
A well-calibrated model produces smoother output distribution, which leads to better generalization and feature learning. In this experiment, we investigate the behavior of a better-calibrated model in the TD validation. As LS can prevent the over-confident prediction and improve probability calibration \cite{muller2019does}, we observe the TD data prediction performance of our approach trained between w and w/o LS. Table \ref{table:calibration_vs_da} demonstrates the comparative scores with the TD prediction and calibration errors. The results indicate the superiority of the calibrated model on the DA task.

\subsection{Qualitative Analysis}
We visualize some examples predicted by our proposed model for the source and target domain, as shown in Fig. \ref{fig:Results visualization}. Sometimes there are redundant words at the end of the predicted sentence, which may be caused by exposure bias \cite{ranzato2015sequence} and word-level training by using the traditional CE loss \cite{ranzato2015sequence}. This problem will be solved by introducing a Reinforcement Learning training technique to achieve sequence-level training \cite{rennie2017self} and fine-tune the sequence generation. Overall, the generated sentences can describe the content of the surgical images accurately.

\section{Ablation Study}

\subsection{LS Feature Extraction}
\begin{figure}[!h]
\centering
\includegraphics[width=1\linewidth]{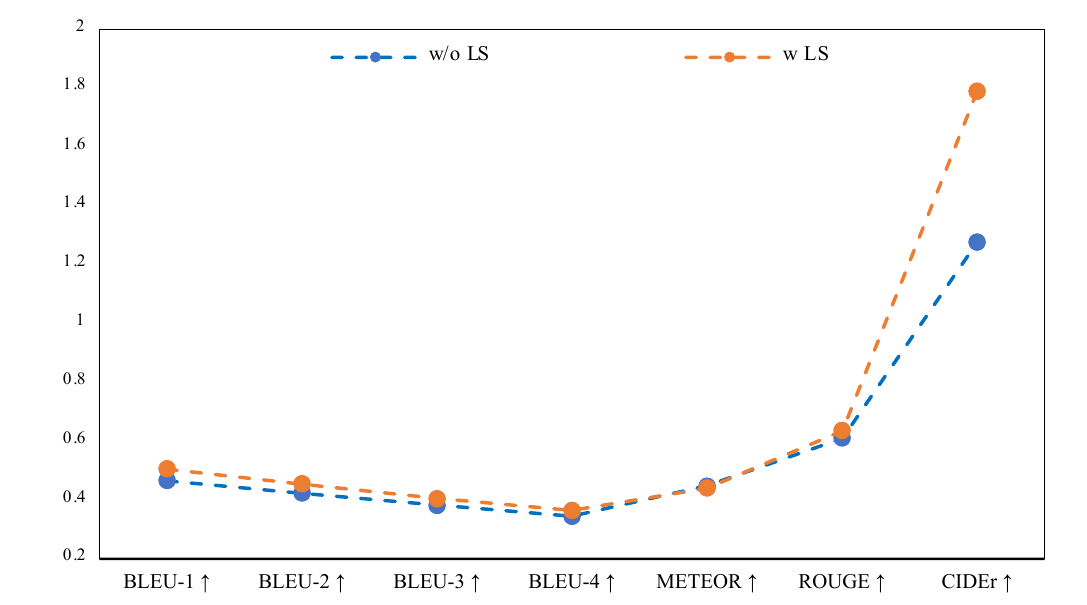}
\caption{Caption prediction performance w and w/o LS feature extraction using ResNet18. LS extracted feature boosts the prediction performance.}
\label{fig:caption_with_ls_features}
\end{figure}

Fig. \ref{fig:tsne_ls_features} is illustrated the better feature representation extracted from ResNet18 classification model trained with LS. In this section, we present the caption prediction models' performance, train w, and w/o LS extracted features. Fig. \ref{fig:caption_with_ls_features} demonstrates the significant prediction improvement using LS extracted feature.

\begin{table}[!h]
\centering
\caption{Ablation study of the proposed model while integrating LS and GRL for the SD and the TD (UDA) experiments.}
\label{table:module_ab}
\begin{tabular}{c|c|c|c|c|c|c}
\hline
\multicolumn{3}{c|}{\textbf{Modules}} &\multicolumn{2}{c|}{\textbf{SD}} &\multicolumn{2}{c}{\textbf{TD (UDA)}} \\ \hline

\textbf{Base} & \textbf{LS} & \textbf{GRL} & BLEU-1 &CIDEr & BLEU-1 &CIDEr \\ \hline
\cmark & \cmark & \cmark &0.5228   &2.2598  &0.2493 &0.1517\\ \hline
\cmark & \cmark & \xmark &0.5161   &2.3406 &0.2345  &0.1314\\ \hline
\cmark & \xmark & \xmark  &0.4659  &1.2752  &0.2302  &0.1305 \\ \hline

\end{tabular}
\label{fig:ablation_grl_dc}
\end{table}

\subsection{GRL Domain Classifier}
Table \ref{fig:ablation_grl_dc} shows the effect of each proposed technique in this work. GRL domain classifier boosts the model prediction of both SD and TD datasets for most of the metrics. However, LS produces a significant enhancement in SD prediction. Table \ref{table:outputs_of_domain classifier} demonstrates that the extra class improves the model performance by making the model harder to be fooled.

\begin{table}[!h]
\centering
\caption{Two vs. Three outputs of GRL Domain Classifier}
\begin{tabular}{c|c|c|c}
\hline
\multirow{2}{*}{\begin{tabular}[c]{@{}c@{}}GRL \\ \textbf{Domain Classifier}\end{tabular}} & \multicolumn{3}{c}{\textbf{SD}}  \\ \cline{2-4} 
                                                                                  & BLEU-1 & METOR  & CIDEr  \\ \hline
Two classes (SD, TD)                                                                         & 0.5231 & 0.4566 & 2.1547 \\ \hline
Three classes (SD, TD, Extra)                                                                        & 0.5228 & 0.4567 & 2.2598 \\ \hline
\end{tabular}
\label{table:outputs_of_domain classifier}
\end{table}

\section{Discussion and Conclusion}
We present a multi-layer encoder-decoder transformer-like model incorporating the gradient reversal adversarial learning and LS to generate captions for the SD and the TD surgical images, which can describe the semantic relationship between instruments and surgical ROI. We observe that a well-calibrated model can be beneficial for domain adaption tasks. The experimental results prove that the proposed model can perform better than the base model in the SD and the TD. When these sentences are stacked, a medical report for the surgical procedures can be obtained. The limitation is the small TD dataset. The performance can be further improved if we can use a larger dataset. Future work can be focused on integrating temporal information in surgical report generation. 

\section{Acknowledgment}
This work was supported by the National Key Research and Development Program, The Ministry of Science and Technology (MOST) of China (No. 2018YFB1307700), Singapore Academic Research Fund under Grant R397000353114 and TAP Grant R-397-000-350-118.

\bibliographystyle{IEEEtran}
\bibliography{mybib}

\end{document}